\documentclass{article}
\usepackage{PRIMEarxiv}
\usepackage[utf8]{inputenc} 
\usepackage[T1]{fontenc}    
\usepackage{hyperref}       
\usepackage{url}            
\usepackage{booktabs}       
\usepackage{amsfonts}       
\usepackage{nicefrac}       
\usepackage{microtype}      
\usepackage{lipsum}
\usepackage{fancyhdr}       
\usepackage{graphicx}       
\graphicspath{{media/}}     
\usepackage{footnote}
\usepackage{subfigure}
\usepackage{multirow}
\usepackage{amsmath}
\usepackage{amsthm}
\usepackage{algorithm}
\usepackage{algorithmic}

\usepackage{subfigure}
\usepackage{multirow}

\usepackage{times}
\usepackage{soul}
\usepackage{url}
\usepackage[small]{caption}
\usepackage{graphicx}
\usepackage{booktabs}
\usepackage[switch]{lineno}
\usepackage{amssymb}
\usepackage{xcolor}
\usepackage{marvosym}
\usepackage{ifsym}

\title{DeLELSTM: Decomposition-based Linear Explainable LSTM to Capture Instantaneous and Long-term Effects in Time Series
}

\author{
  Chaoqun Wang, Yijun Li, Xiangqian Sun, Qi Wu{\thanks{Corresponding author.}}\\
  School of Data Science\\
  City University of Hong Kong \\
  \texttt{\{cqwang5-c, yijunli5-c, xqsun4-c\}my.cityu.edu.hk,} \\
   \texttt{\{qiwu55@cityu.edu.hk\}} \\
   \And
  Dongdong Wang, Zhixiang Huang \\
  JD Digits \\
  \texttt{\{wangdongdong9, huangzhixiang\}jd.com} \\
}

\begin{document}

\maketitle

\begin{abstract}
Time series forecasting is prevalent in various real-world applications. 
Despite the promising results of deep learning models in time series 
forecasting, especially the Recurrent Neural Networks (RNNs), the explanations 
of time series models, which are critical in high-stakes applications, have 
received little attention. In this paper, we propose a Decomposition-based 
Linear Explainable LSTM (DeLELSTM) to improve the interpretability of LSTM. 
Conventionally, the interpretability of RNNs only concentrates on the variable 
importance and time importance. We additionally distinguish between the 
instantaneous influence of new coming data and the long-term effects of 
historical 
data. Specifically, DeLELSTM consists of two components, i.e., standard LSTM 
and tensorized LSTM. The tensorized LSTM assigns each variable with a unique 
hidden state making up a matrix $\mathbf{h}_t$, 
and the standard LSTM models 
all the 
variables with a shared hidden state $\mathbf{H}_t$. 
By decomposing the $\mathbf{H}_t$ 
into the linear combination of past 
information $\mathbf{h}_{t-1}$ and the fresh information $\mathbf{h}_{t}-\mathbf{h}_{t-1}$, 
we can get the 
instantaneous influence and the long-term effect of each variable. In addition, 
the advantage of linear regression also makes the explanation transparent and 
clear. We demonstrate the effectiveness and interpretability of DeLELSTM on 
three empirical datasets. Extensive experiments show that the proposed method 
achieves competitive performance against the baseline methods and provides a 
reliable explanation relative to domain knowledge.
\end{abstract}

\section{Introduction}
Time series forecasting is ubiquitous across a broad range of applications, 
including finance~\cite{wu2013dynamic}, 
meteorology~\cite{chakraborty2012fine}, energy consumption~\cite{wang2020lstm}, 
and medical 
health~\cite{zhang2019attain}. Deep neural networks have been successfully 
developed for time series forecasting tasks. Among them, Recurrent Neural 
Networks (RNNs) and its variant, the Long-Short Term Memory (LSTM) 
\cite{hochreiter1997long} and the 
Gated Recurrent Units (GRU) \cite{cho2014learning}, are widely used networks 
for 
handling sequential data and have been proven to be powerful tools 
\cite{guo2019exploring}. 

Despite the promising results of RNNs in time series forecasting, using RNNs to 
modelling time series lack interpretability, which is critical in high-stakes 
applications like finance and healthcare. For example, in the medical field, 
e.g., predicting the mortality rate after patients enter the ICU, it is crucial 
for clinicians to understand how models output a specific prediction and which 
indicators are useful. Such explanations can aid reliable decision-making for 
clinicians and increase trust in models’ predictions. 

Although much recent work has been done on explainability in the computer 
vision and natural language processing 
\cite{masoomi2021explanations,mohankumar2020towards,tsang2020does}, this 
problem has been overlooked  in the 
case of time series forecasting \cite{tonekaboni2020went,rojat2021explainable,hsieh2021explainable}. The time series is special because of its dynamic 
nature, which causes multivariable patterns to change over time and makes it 
more difficult to build explainable models. 

There are two main challenges in explaining time series forecasting models. 
First, each variable has a different impact on the target series, and the 
effects of variables on the target series are dynamic over time. Therefore, 
capturing 
the different dynamic impacts of each variable and distinguishing the 
contribution of each variable to the prediction is difficult for explaining the 
forecasting model. Second, the serial dependencies for each variable are 
heterogeneous. When predicting target series, the long-term effects of some 
variables play a decisive role, while the instantaneous effect of other 
variables is more important. For example, in the 
financial market, investors tend to utilize multivariate time series, such as 
the stock index and related stocks, to forecast the stock prices at the next 
time point. If the instantaneous influence and the long-term 
effects of each feature can be distinguished, investors can focus on more 
significant information and make appropriate investment decisions. As a result, 
we should solve the challenge of distinguishing the long-term effect and 
the instantaneous influence of each 
variable. 

Existing explainable RNNs primarily use the attention 
mechanism on the hidden states to get the important variables 
\cite{tonekaboni2020went,guo2019exploring}. The hidden 
states involve information from previous steps and the new inputs of all 
variables, but none of these methods explicitly model the immediate impact of 
fresh information and the long-term effects of historical data. In addition, 
whether the attention mechanism can be directly applied to model interpretation 
is still controversial \cite{sun2021interpreting,serrano2019attention}. 
Moreover, most of the explainable studies just make predictions once based on 
historical data and ignore the importance of real-time series 
forecasting, which predicts at each time point. For example, predicting 
electricity consumption at each hour is crucial for electric power providers to 
maximize resource utilization and cut costs. 

To address these challenges, we propose DeLELSTM: Decomposed-based Linear 
Explainable LSTM, which simultaneously: (i) identifies the instantaneous 
influence 
and long-term effects of each variable; (ii) leverages the linearity of linear 
regression to make interpretation transparent and straightforward. 
Specifically, DeLELSTM involves two components. One is standard LSTM, in which 
the hidden state $\mathbf{H}_t$ encapsulates information of all variables until 
time 
$t$. The other is 
tensorized LSTM \cite{he2017wider}, where the hidden state $\mathbf{h}_t$ is a 
matrice and 
each row of the hidden state only encodes information exclusively from a 
certain variable of the input. To capture the instantaneous influence and 
long-term effects of each variable, the linear combination of the hidden state 
at time $t-1$ and the dynamic change of the hidden state at time $t$ of each 
variable is used to approximate the $\mathbf{H}_t$. The linearity of the linear 
regression can make the explanation transparent and clear. In addition, given 
the 
significance 
of real-time forecasting in time series, we aim to build an explainable model 
for real-time series forecasting. 

The major contributions of this work are as follows: 
\begin{itemize}
	\item We consider the instantaneous influence and long-term effects of each 
	variable, which commonly exists in time series data.
	\item We take advantage of the explanation of the linear regression model 
	to 
	make a clear and transparent explanation.
	\item Extensive experiments with three benchmark datasets show that 
	DeLELSTM can achieve competitive performance and provide transparent 
	explanations relative to domain knowledge.
\end{itemize}

The rest of the paper is organized as follows. Section 2 reviews the related 
work. 
Section 3 introduces the problem definition and details our proposed 
framework. 
Then we evaluate our method by comparing it with several baselines in Section 
4 
and conclude in Section 5.

\section{Related Work}
In recent years, different classes of approaches have been proposed to explain 
time 
series forecasting models, especially RNNs. One general approach is post-hoc 
analysis, which explains the models by evaluating the importance of each 
variable in the 
predictions. The other is ante-hoc methods, which incorporate interpretability 
directly into their structures. 

Post-hoc methods frequently adopt variable-level attribution interpretations, 
also known as salience maps \cite{ding2022explainability}. These methods assign 
a relevance score to each variable, indicating how sensitive a variable is to 
the 
output. Gradient-based and perturbation-based methods are the two main types of 
attribution methods. Gradient-based methods examine the characteristics to 
which output was most sensitive 
\cite{ancona2017towards,shrikumar2017learning,sundararajan2017axiomatic,yang2018explaining}. In perturbation-based 
approaches, the variable importance is obtained by perturbing the variables 
with 
mean value or random uniform noise, running a forward pass on the new put, and 
comparing the difference to the original output 
\cite{dabkowski2017real,fong2017interpretable}. However, these post-hoc 
explainable models have been 
criticized for failing to capture the sequential dependencies and clarifying 
how 
the underlying model arrived at a specific prediction 
\cite{tonekaboni2020went,rigotti2021attention}. 

On the other hand, ante-hoc models can provide intrinsic explanations by 
building self-explanatory systems. Among ante-hoc models, attention-based 
interpretable models are widely used for explaining RNNs. The parameters 
of these models, known as attention weights, are utilized to explain how the 
models behave over time. For example, ~\cite{choi2016retain} 
proposed RETAIN, an explainable model based on a two-level attention mechanism. 
Two sets of attention scores are used to identify relevant clinical variables 
and influential hospital visits, respectively.  
~\cite{guo2019exploring} proposed a mixture attention framework to get 
variable importance and temporal importance. However, the focus of the current 
attention mechanism is on hidden states, which encode information from both 
past and 
new observations. Differentiating between the immediate impact of new 
information and the long-term effects of historical data can be difficult. In 
addition, attention’s interpretability is still debatable 
~\cite{sun2021interpreting}, and other studies found 
that attention patterns cannot reliably provide transparent and reliable 
explanations ~\cite{jain2019attention}. 

In this paper, we propose a decomposition-based method to decompose hidden 
states at each time step into two components, representing instantaneous 
information and historical information, respectively. In addition, the linear 
regression model is utilized to build a transparent and faithful explainable 
model for time series forecasting. 

\section{The Proposed Framework - DeLELSTM}
\subsection{Problem Definition}
Let $\textbf{X}\in \mathbb{R}^{D\times T}$ be a sample of a multivariate 
time-series 
data where $D$ is the number of variables with $T$ observations over time. 
Further, $\mathbf{x}_t \in \mathbb{R}^D$ is the set of all observations at time 
$t \in 
[1,\cdots,T]$, denoted by the vector 
$[x_{t}^1,x_t^2,\cdots,x_t^D]$ and 
$\mathbf{X}_{1:t} \in \mathbb{R}^{D\times t}$ is the matrix 
$[\mathbf{x}_{1};\mathbf{x}_{2};\cdots;\mathbf{x}_{t}]$, representing the 
observations until $t$.

Let $\textbf{Y}\in \mathbb{R}^{T}$ be the target time series of length $T$. 
Noted, the target series can be one of the multivariate time series 
$\textbf{X}$ or not.
Given 
$\textbf{X}_{1:t}\in \mathbb{R}^{D\times t}$, we aim to learn a function 
$\mathcal{F} $ to predict the next value 
of 
the target series, namely, $y_{t+1}=\mathcal{F}(\textbf{X}_{1:t})$

\subsection{Network Architecture: DeLELSTM}
This subsection first describes the framework of our proposed model 
DeLELSTM, then follows with the details of the model and the process of 
obtaining the interpretation.

\begin{figure*}[ht]
	\centering
	\includegraphics[scale= 0.6]{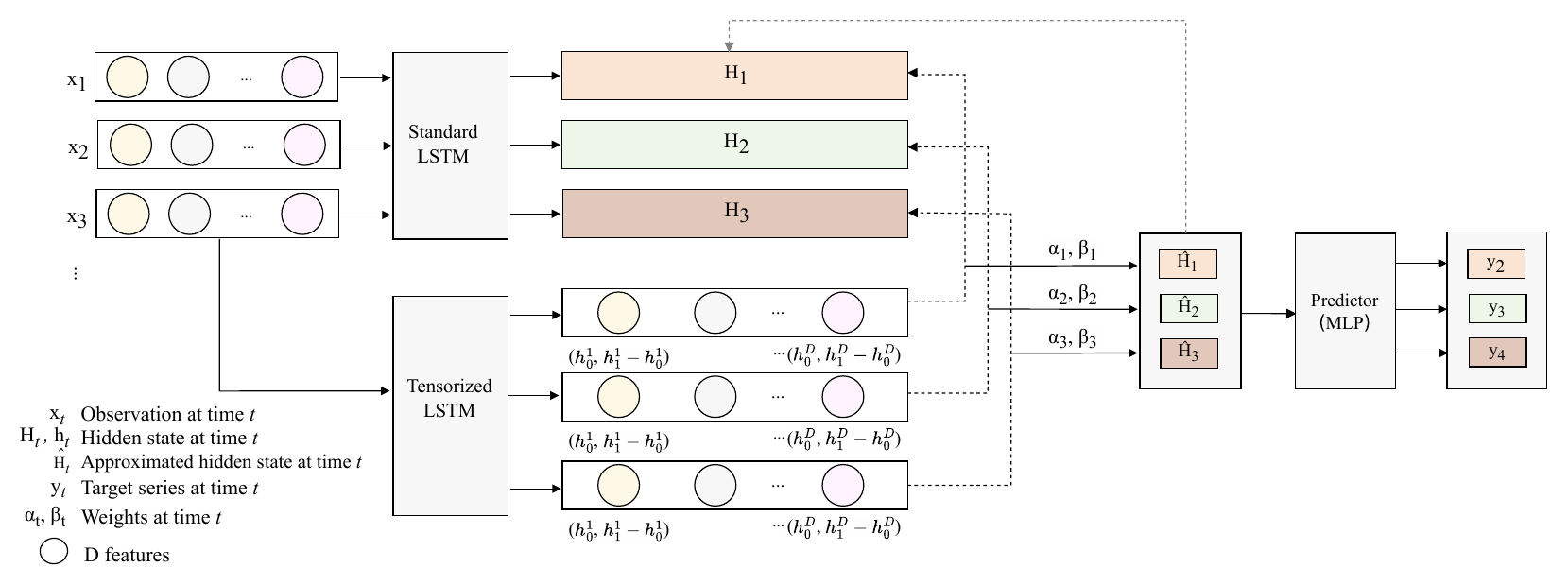}
	\caption{Proposed DeLELSTM model Framework.}
	\label{framework}
\end{figure*}

DeLELSTM 
consists of two components (i) a standard LSTM, in which $\mathbf{x}_t$ is the 
input at 
time $t$ and $\mathbf{H}_t$ is the hidden state at $t$. The hidden state 
$\mathbf{H}_t$ 
incorporates new 
information 
and past information of all variables; (ii) a tensorized LSTM, in which the 
hidden state $\mathbf{h}_t$ is a matrix and each row of the hidden 
state  $\mathbf{h}_t$ only contains 
information taken from one particular input variable.  $\mathbf{h}_{t-1}$ 
stands for the 
information from the past, and $\mathbf{h}_t-\mathbf{h}_{t-1}$ is the dynamic 
change brought by 
the 
new observations at time $t$. Here, we define $\Delta 
\mathbf{h}_t=\mathbf{h}_t-\mathbf{h}_{t-1}$. To capture both the immediate 
influence and the 
long-term effect of each variable, we propose to approximate $\mathbf{H}_t$ as 
a linear 
combination of $\mathbf{h}_{t-1}$  and the dynamic change 
 $\Delta 
\mathbf{h}_t$, 
thereby 
allowing the separation of output states into contributions from instantaneous 
influence and 
long-term effects of each variable. Finally, the approximated hidden state 
$\hat{\mathbf{H}}_t$ is used to 
predict the $y_{t+1}$ and the computation of the hidden state in the next step 
$\mathbf{H}_{t+1}$. Figure \ref{framework} shows the architecture of the 
proposed 
DELeLSTM 
framework. 

The architecture of DeLELSTM is made to learn a representation of the 
multivariate 
time series data that suffices for accurate real-time prediction and offers a 
transparent explanation of each variable’s long-term effect and instantaneous 
influence. We now proceed to illustrate each module of DeLELSTM in more detail. 

\subsubsection{Standard LSTM}
\normalsize
The standard LSTM is shown in Equation \eqref{LSTM}. 
\begin{equation}\label{LSTM}
	\begin{aligned}
	&\mathbf{I}_t=\sigma({U}_{i}\mathbf{H}_{t-1}+{W}_{i}\mathbf{x}_t+\mathbf{B}_i) \\
	&\mathbf{F}_t=\sigma({U}_{f}\mathbf{H}_{t-1}+{W}_{f}\mathbf{x}_t+\mathbf{B}_f) \\
	&\mathbf{O}_t=\sigma({U}_{o}\mathbf{H}_{t-1}+{W}_{o}\mathbf{x}_t+\mathbf{B}_o) \\
	&\mathbf{\tilde{C}}_t=\tanh({U}_{c}\mathbf{H_{t-1}}+{W}_{c}\mathbf{x}_t+\mathbf{B}_c) \\
	&\mathbf{C}_t=\mathbf{F}_t\odot \mathbf{C}_{t-1}+\mathbf{I}_t\odot\mathbf{\tilde{C}}_t\\
	&\mathbf{H}_t=\mathbf{O}_t \odot \tanh(\mathbf{C}_t)	
    \end{aligned}
\end{equation}

where $\mathbf{x}_t\in \mathbb{R}^D$ is the new observations of $D$ variables 
at time 
$t$, $\mathbf{H}_t \in \mathbb{R}^M$ is the hidden state at time $t$ which 
encodes 
information from all input variables until time $t$. $M$ is the dimension of 
the 
hidden state. $\odot$ is the 
elementwise product. 
\subsubsection{Tensorized LSTM}
The tensorized LSTM is used to explain and approximate the hidden state 
$\mathbf{H}_t$ 
and identify the 
contribution of each variable. Here, the tensorized LSTM can be considered as a 
set of parallel LSTMs, each of which processes a single variable series. The 
hidden state of the tensorized LSTM is a matrix, and each row of the hidden 
state incorporates information only from a particular variable. Equation 
\eqref{TensorLSTM} shows the tensorized LSTM.
\begin{equation}\label{TensorLSTM}
	    \begin{aligned}
	    &\mathbf{i}_t=\sigma(\mathcal{U}_i\circledast
	    \mathbf{h}_{t-1}+\mathcal{W}_i \circledast
	    \mathbf{x}_{t}+\mathbf{b_i})\\
	    &\mathbf{f}_t=\sigma(\mathcal{U}_f\circledast
	    \mathbf{h}_{t-1}+\mathcal{W}_f \circledast
	    \mathbf{x}_{t}+\mathbf{b_f})\\
	    &\mathbf{o}_t=\sigma(\mathcal{U}_o\circledast
	    \mathbf{h}_{t-1}+\mathcal{W}_o \circledast
	    \mathbf{x}_{t}+\mathbf{b_o})\\
	    &\mathbf{\tilde{c}}_t=\tanh(\mathcal{U}_{c}\circledast\mathbf{h}_{t-1}+\mathcal{W}_c\circledast
    \mathbf{x}_t+\mathbf{b}_c)\\
        &\mathbf{c}_t=\mathbf{f}_t\odot 
        \mathbf{c}_{t-1}+\mathbf{i}_t\odot\mathbf{\tilde{c}}_t\\
		&\mathbf{h}_t=\mathbf{o}_t \odot \tanh(\mathbf{c}_t)
		\end{aligned}\\
\end{equation}
where $\mathbf{h}_t=[\mathbf{h}_t^1,\cdots,\mathbf{h}_t^D]^T,\mathbf{h}_t\in 
\mathbb{R}^{D \times M}$,$\mathbf{h}_t^d\in 
\mathbb{R}^M$. The $\mathbf{h}_t^d$ is the hidden state vector specific to the 
$d$-th 
input variable at time $t$, and the $\mathbf{h}_t$ is the hidden state of all 
variables 
at time $t$.
$\mathcal{U}_{c}=[\mathbf{u}_c^1,\cdots,\mathbf{u}_c^D], ~\mathcal{U}_{c} \in 
\mathbb{R}^{D\times M \times M}, ~\mathbf{u}_c^d \in \mathbb{R}^{M \times M}$, 
is the 
hidden-to-hidden transition. 
$\mathcal{W}_{c}=[\mathbf{w}_c^1,\cdots,\mathbf{w}_c^D], ~\mathcal{W}_{c} \in 
\mathbb{R}^{D\times M \times 1}, ~\mathbf{w}_c^d \in \mathbb{R}^{M \times 1}$, 
is the 
input-to-hidden transition. ($\mathcal{U}_i,\mathcal{U}_f,\mathcal{U}_o$), (
$\mathcal{W}_i,\mathcal{W}_f,\mathcal{W}_o$) have the same shapes as 
$\mathcal{U}_{c},\mathcal{W}_{c}$, respectively. $\circledast$ is the 
tensor-dot operation, which is defined as the product of two tensors along the 
$D$ axis, e.g., 
$\mathcal{U}_{c}\circledast
\mathbf{h}_{t-1}=[\mathbf{u}_c^1\mathbf{h}_{t-1}^1,\cdots,\mathbf{u}_c^D\mathbf{h}_{t-1}^D]^T$,
$\mathbf{u}_c^d\mathbf{h}_{t-1}^d \in \mathbb{R}^M$. Such a design can 
guarantee that gates and 
memory cells are also matrices, and each row of these matrices only captures 
the 
information from a single variable.  
\subsubsection{Decomposition}
$\mathbf{H}_t$ involves the information from previous steps and new input of 
all 
variables, so it is challenging to measure the influence of each variable, 
including the long-term effect from $\mathbf{H}_{t-1}$ and the instantaneous 
impact from 
$\mathbf{x}_t$.

Theoretically, each row of $\mathbf{h}_{t-1}$ corresponds to the information 
belonging 
to 
a 
particular variable until time $t-1$. Each row of 
$\Delta \mathbf{h}_{t}$ represents 
a 
single variable’s new information of time $t$, 
We propose to decompose 
$\mathbf{H}_t$ 
as a 
linear combination of $\mathbf{h}_{t-1}$ and $\Delta \mathbf{h}_{t}$, 
so that we can obtain each 
variable’s 
instantaneous influence and long-term impact.

Specifically, we approximate $\mathbf{H}_t$ using Equation 
\eqref{Approximation}:
\begin{equation}\label{Approximation}
	\begin{aligned}
		\mathbf{H}_t \approx \sum_{i=1}^{D} (\alpha_{t}^{i} \mathbf{h}_{t-1}^{i}+\beta_t^i 
		(\mathbf{h}_t^i-\mathbf{h}_{t-1}^i))
		\\=\mathbf{h}_{t-1}^T  \boldsymbol{\alpha}_t+(\mathbf{h}_{t}-\mathbf{h}_{t-1})^T  \boldsymbol{\beta}_t
	\end{aligned}
\end{equation}
This approximation is minimised following a least squares criterion, which has 
a well-known 
solution, $\hat{\boldsymbol{\alpha}}_t$, $\hat{\boldsymbol{\beta}}_t$, and 
corresponding optimal approximation $\hat{\mathbf{H}}_t$. Here, we use 
$\hat{\mathbf{H}}_t$ to 
get the  prediction of target $y_{t+1}$ and compute the hidden state 
$\mathbf{H}_{t+1}$.

\subsection{Learning to Interpret}
To get the immediate impact and the long-term effect of each variable, we 
derive 
significance from the magnitude of linear approximation weights rather than 
only focusing on the largest values. We first take the absolute value of 
$\hat{\boldsymbol{\alpha}}_t$, $\hat{\boldsymbol{\beta}}_t$ and get $\tilde{\boldsymbol{\alpha}}_{t}$, $\tilde{\boldsymbol{\beta}}_t$ 
after normalizing weights at each time step. 
Then, we propose several measures to interpret the prediction model for 
real-time series forecasting. 
\newtheorem{definition}{Definition} 
\begin{definition}
	The instantaneous importance of the d-th variable at time t, $In_t^d$, 
	is defined as the Equation \eqref{instantaneous}, the ratio of 
	$\tilde{\beta}_t^{d}$ and  $(\tilde{\alpha}_t^{d}+ 
	\tilde{\beta}_t^{d})$. Accordingly, the long-term effect of the d-th
	variable at time t is $1-In_t^d$.
\begin{equation}\label{instantaneous}
	In_t^d=\frac{\tilde{\beta}_t^{d}}{(\tilde{\alpha}_t^{d}+ 
		\tilde{\beta}_t^{d})}
\end{equation}
\end{definition}
$In_t^d$ can help us figure out which is more important: past information until 
time $t-1$ versus current data at time $t$ of the $d$-th variable. If $In_t^d$ 
is close to 1, we can know that the past information of the variable $d$ has 
little impact on the prediction; in other words, it has no long-term effect. 

\begin{definition}
	The global importance of the d-th variable at time T, $Gl_T^d$,
	is defined as the Equation \eqref{global}, which considers both the 
	immediate impact and long-term impact of the d-th variable.
\begin{equation}\label{global}
	Gl_T^d=\frac{1}{T}\sum_{t=1}^{T}\sqrt{(\tilde{\alpha}_t^{d}{}^2
		+\tilde{\beta}_t^{d}{}^2)}
\end{equation}
\end{definition}
The global importance of each variable can help us identify the important 
variables in general. Further, we can also get the dynamic shift in a 
variable's weight by considering $\sqrt{(\tilde{\alpha}_t^{d}{}^2
	+\tilde{\beta}_t^{d}{}^2)}$ over time, which we use to provide further model 
interpretability.

Algorithm \ref{alg:algorithm} summarizes the proposed procedure.

\begin{algorithm}[tb]
	\caption{The Training Process of DeLELSTM}
	\label{alg:algorithm}
	\textbf{Input}:Time series $\mathbf{X}_{1:T}$, where $T$ is the max time 
	length;\\
	\textbf{Set}: $\mathbf{H}_t=\mathbf{H}_0$, $\mathbf{h}_t=\mathbf{h}_0$ \\
	\textbf{Output}: $\boldsymbol{\alpha}_{1:T}$, $\boldsymbol{\beta}_{1:T}$, 
	$\mathbf{y}_{2:T+1}$
	\begin{algorithmic}[1] 
		\FOR{$t=1$ to $T$}
		\STATE $\mathbf{H}_t$=standard 
		LSTM($\mathbf{x}_t$, $\mathbf{H}_{t-1}$);
		\STATE  $\mathbf{h}_t$=Tensorized 
		LSTM($\mathbf{x}_t$, $\mathbf{h}_{t-1}$);\\
		\STATE $\Delta \mathbf{h}_{t}=\mathbf{h}_t-\mathbf{h}_{t-1}$;\\
		\textcolor{blue}{\% Compute the dynamic change of hidden state 
			$\mathbf{h}_t$ 
			from $t-1$ to $t$}\\ 
		\STATE $(\boldsymbol{\alpha}_t,\boldsymbol{\beta}_t)$=Least 
		Square$(\mathbf{H}_t, \mathbf{h}_{t-1}, \Delta \mathbf{h}_t)$;\\
		\textcolor{blue}{\% Compute the optimal solution of 
			$(\boldsymbol{\alpha}_t,\boldsymbol{\beta}_t)$ using Leaset Square 
	}\\
		\STATE 
		$\hat{\mathbf{H}}_t=(\mathbf{h}_{t-1})^T\boldsymbol{\alpha}_t+
		(\Delta \mathbf{h}_t)^T
		\boldsymbol{\beta}_t$\\
		\textcolor{blue}{\% Compute the approximated hidden state 
		$\hat{\mathbf{H}}_t$}\\
		\STATE $\hat{y}_{t+1}$=MLP($\hat{\mathbf{H}}_t)$
		\ENDFOR
		\STATE \textbf{return} $\boldsymbol{\alpha}_{1:T}$, 
		$\boldsymbol{\beta}_{1:T}$, 
		$\mathbf{y}_{2:T+1}$
	\end{algorithmic}
\end{algorithm}
\section{Experiment}
In this section, we describe our experiments to evaluate the prediction performance and the interpretation ability of our proposed model \footnote{Code and supplementary is available in the repository:  https://github.com/wangcq01/DeLELSTM. \label{supp}}.
\subsection{Datasets}
We used three publicly available real-world multivariate time series datasets 
covering meteorology, energy, and finance fields.

\textbf{PM2.5} \cite{liang2015assessing}: It contains hourly PM2.5 data and the 
associated meteorological measurements in Beijing from 2010.1.1 to 2014.12.31. 
PM2.5 is the target series. Aside from PM2.5 values, the meteorological 
variables include dew point, temperature, pressure, wind direction, wind speed, 
hours of snow, and hours of rain. Given the measurements, the task is to 
forecast PM2.5 each hour within a day. For example, using the data 
before 2:00 predicts PM2.5 at 2:00; using the data before 3:00 predict  its 
value at 3:00, 
etc.

\textbf{Electricity} \cite{gao2022explainable}: It records the time series of 
electricity consumption in the US, from 2017.10.11 to 2020.6.24. The 
consumption is 
chosen as the target series sampled hourly. The other 15 time series are 
exogenous factors, including max temperature, min temperature, visibility, etc. 
Similarly to PM2.5, our task is also to forecast consumption each 
hour of the day.

\textbf{Exchange} \cite{lai2018modeling}: It is the collection of the daily 
exchange rates of eight foreign countries, including Australia, British, 
Canada, Switzerland, China, Japan, New Zealand, and Singapore, ranging from 
1990 to 2016. We consider the time series of 30 days as a sample for this task. 
The Singapore exchange is taken as the target series, and we aim to predict the 
exchange rate of Singapore each day of a month. 
\subsection{Baseline Methods and Evaluation Setup}
We evaluate the prediction performance of DeLELSTM with several baseline 
models as follows:

\textbf{LSTM}:\cite{hochreiter1997long}: LSTM network with one hidden layer is 
used to learn an encoding from multivariate time series data and make 
predictions at each time step. The prediction performance is also used as the 
basis for other models.

\textbf{RETAIN}:\cite{choi2016retain}: RETAIN is a two-level neural attention 
model for sequential data that can recognize the influential events and 
relevant characteristics within these events.

\textbf{IMV-LSTM}:\cite{guo2019exploring}:IMV-LSTM explores the structure of 
LSTM network to learn variable-wise hidden states and separate the contribution 
of variables to the prediction. With hidden states, a mixture attention 
mechanism is explored to model the generative process of the target. It has two 
realizations of IMV-LSTM, i.e., \textbf{IMV-Full} and \textbf{IMV-Tensor}. We 
consider both versions.

We implemented the proposed model and deep learning baseline methods with 
Pytorch. We used Adam\cite{kingma2014adam} optimizer. We conduct the grid 
search to select optimal parameters. The batch size is selected in $\{32, 64, 
128\}$. Learning rate is searched in $\{0.05, 0.01, 0.001\}$. The size of the 
hidden states is selected in $\{32, 64, 128\}$. We train the models using 
$75\%$ of the samples, and $15\%$ of the samples are for validation. The 
remaining $10\%$ is used as the test set. We repeat the experiment five times 
and report the average performance with standard deviation.

We consider three metrics to measure the prediction performance, i.e., RMSE,  
MAE, and MAPE. RMSE is defined as $RMSE=\sqrt{\sum_{n}(y_n-\hat{y}_n)^2/N}$. 
MAE is 
defined as $MAE=\sum_{n}\|y_n-\hat{y}_n\|/N$. MAPE is defined as 
$MAPE=\sum_{n}(\lvert \hat{y}_n-y_n\rvert/\lvert y_n \rvert)/N \times 100\%$, 
where 
$\hat{y}_n$ is the predicted value, and ${y}_n$ is the true value.
\subsection{Prediction Performance}
We compared the proposed DeLELSTM with four baseline models on real-time 
series forecasting and reported the results in Table \ref{performance}. 
As shown in Table \ref{performance}, 
among the 
attention-based models, the performance of RETAIN is better than IMV-LSTM in 
most cases. The IMV-Full is better than IMV-Tensor. Our proposed model presents comparable performance and obtains the best performance in electricity 
consumption prediction, indicating that our decomposition-based linear 
explainable model can still guarantee the prediction performance, while 
capturing the instantaneous and long-term effects and providing transparent and 
clear interpretation.

\begin{table*}
		\centering
	\begin{tabular}{llccccc}
		\hline
		Dataset                & Metric & LSTM                      & 
		RETAIN                & IMV-Full         & IMV-Tensor       & 
		DeLELSTM         \\
		\hline
		\multirow{3}{*}{Electricity} & RMSE & 5.1049±6.3693     & 2.6685±0.2622     
		& 2.0284±0.1001    & 2.6069±0.9455   & \textbf{1.7247±0.0370}     \\
		& MAE  & 3.9574±5.3274     & 2.0000±0.2136     & 1.4085±0.0899     & 
		1.7526±0.6754   & \textbf{1.1025±0.0128}     \\
		& MAPE & 6.49\% ± 8.84\% & 3.21\% ± 0.36\% & 2.28\% ± 0.17\% & 
		2.74\% ± 1.04\% & \textbf{1.69\%±0.02\%}\\
		\hline
		\multirow{3}{*}{PM2.5} & RMSE   & 26.566 ± 0.152           & 
	\textbf{24.913 ± 0.177} & 28.465 ± 0.264     & 28.866 ± 0.277     & 
		26.612 ± 0.470     \\
		& MAE    & 13.466 ± 0.098              & \textbf{13.402 ± 0.089} & 
		14.087 ± 0.266     & 14.480 ± 0.164     & 13.553 ± 0.168     \\
		& MAPE   & 22.47\%±0.83\% & \textbf{21.23\%±0.42\%}      & 
		25.21\%±1.63\% & 26.67\%±1.64\% & 22.16\%±0.98\%\\
		\hline
			\multirow{3}{*}{Exchange} & RMSE & 0.0026±9.7e-06    & 
			0.0025±9.4e-06     & 0.0026±5.0e-06     & 0.0026±4.6e-06     & 
			\textbf{0.0023±2.7e-05}    \\
			& MAE  & 0.0015±3.8e-06   & \textbf{0.0015±1.0e-05}  & 
			0.0015±5.5e-06     & 0.0015±3.1e-06     & 0.0015±1.5e-05    \\
			& MAPE & 0.23\%±5.5e-06 & \textbf{0.23\% ±1.6e-05} & 
			0.22\% ±3.2e-04 & 0.23\% ±5.7e-06 & 0.23\% ±2.3e-05\\
			\hline		
	\end{tabular}
	\caption{Performance (±standard deviation) of baseline and proposed models}
\label{performance}
\end{table*}
\subsection{Interpretation}
In this subsection, we depict three case studies designed to evaluate the 
effectiveness of DeLELSTM in providing insightful explanations of its 
forecasting. In particular, we qualitatively analyze the immediate and 
long-term impact of each variable identified by the defined measurement 
$ln_t^d$ and $1-ln_t^d$. 
We also report the meaningful variables at each time step and overall according 
to the defined measurement $Gl_T^d$.  
\subsubsection{Case Study I: Electricity Data}
\begin{figure}[tb]
	\centering
	\includegraphics[scale= 0.345]{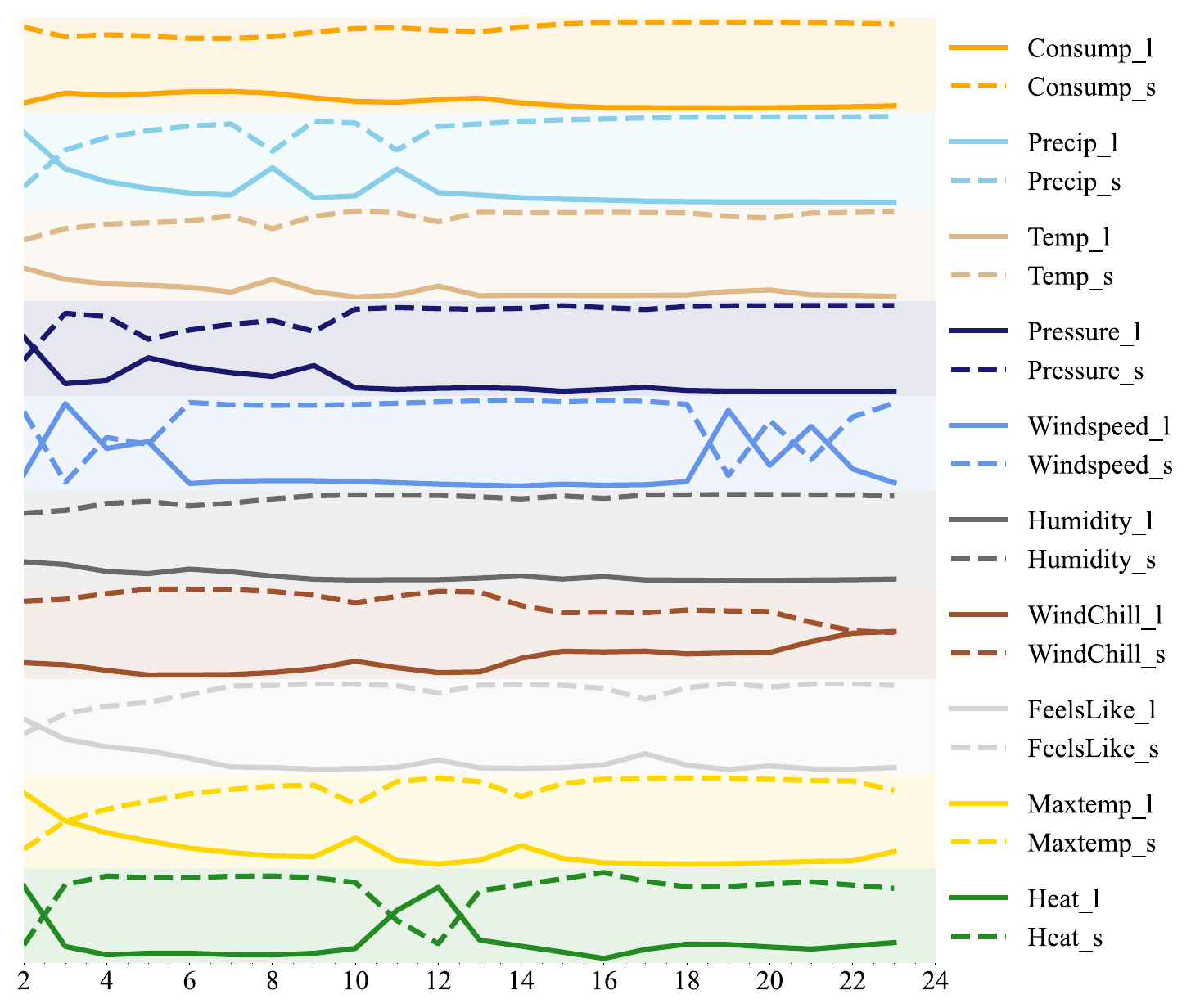}
	\caption{Electricity Long (l)/Short (s) term effects evolving over time (24h)}
	\label{elec_0}
\end{figure}

Predicting real-time electricity consumption accurately is quite important for 
electric power providers, so that they can more precisely manage resources for 
energy generation to maximize resource utilization, cut costs, and advance the 
development of smart grids. Figure \ref{elec_0}  shows how the long-term and instantaneous (short-term) impacts change throughout the day. Here, we depict the change of the top 10 features, ordered by their overall importance for predicting electricity consumption.  We can observe that compared with the short-term effects, the long-term information of most features is vital when people are sleeping and not engaged in other activities, and gradually diminishes importance during the daytime. On the other hand, the long-term effects of WindChill are important for electricity prediction.

\begin{figure}[tb]
	\centering
	\includegraphics[scale= 0.302]{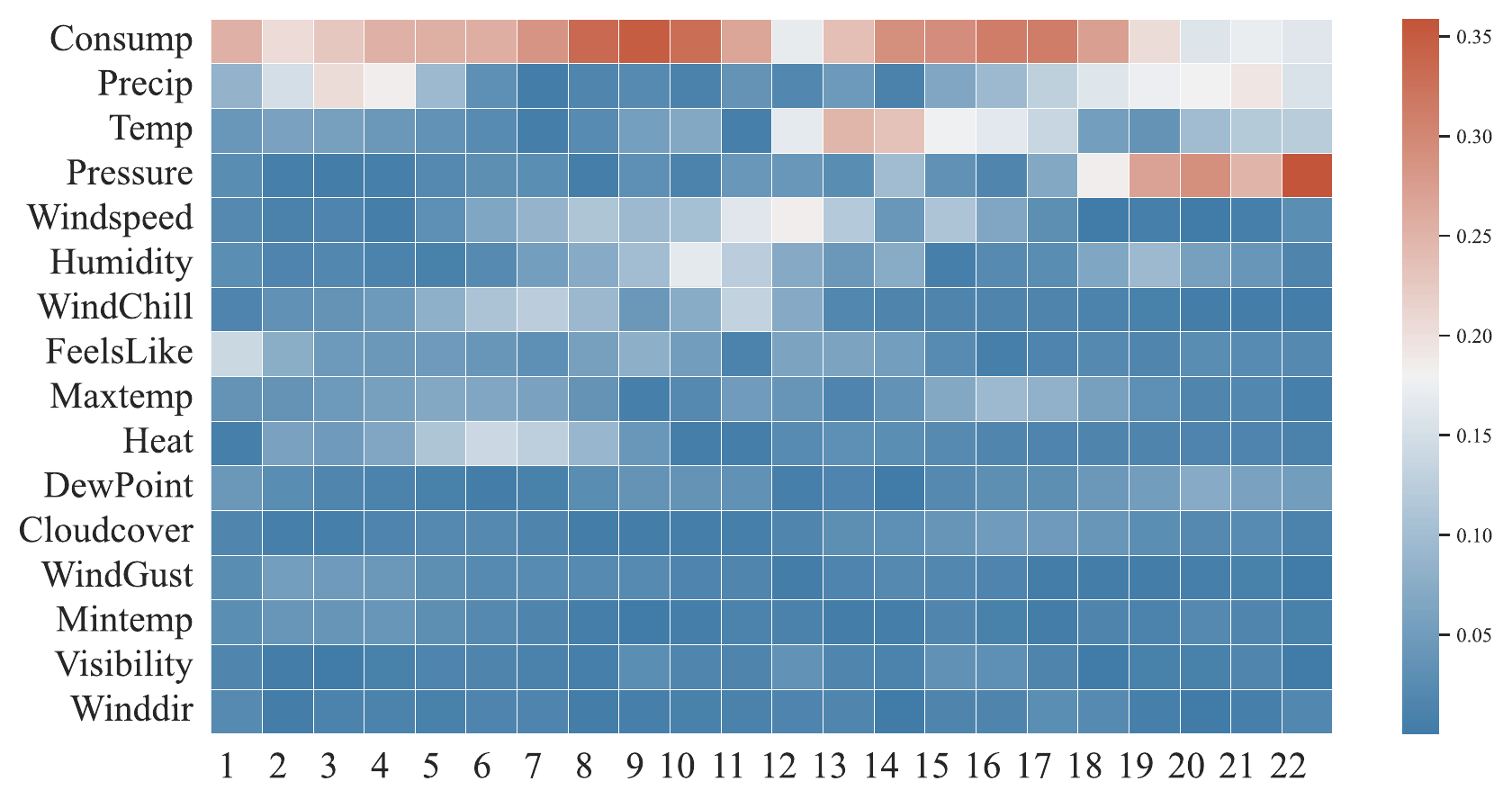}
	\caption{Electricity variable importance.}
	\label{elec_variable}
\end{figure}

Figure \ref{elec_variable} depicts 
the dynamic change of variable importance for the electricity consumption 
forecast. It is seen that the electricity consumption itself has an evident 
auto-correlation and contributes more to the prediction. In addition, as evening approaches, pressure and precipitation become more critical. While the temperature becomes important for electricity consumption at noon, because the temperature at noon tends to be high, people need to switch on the air-conditioner to have a comfortable environment. It is a primary reason that affects electricity consumption at noon. In general, variables consumption, precipitation, temperature, pressure, wind speed and humidity are highly ranked by DeLELSTM.

\subsubsection{Case Study II: PM2.5 Data}
\begin{figure}[ht]
	\centering
	\includegraphics[scale= 
	0.34]{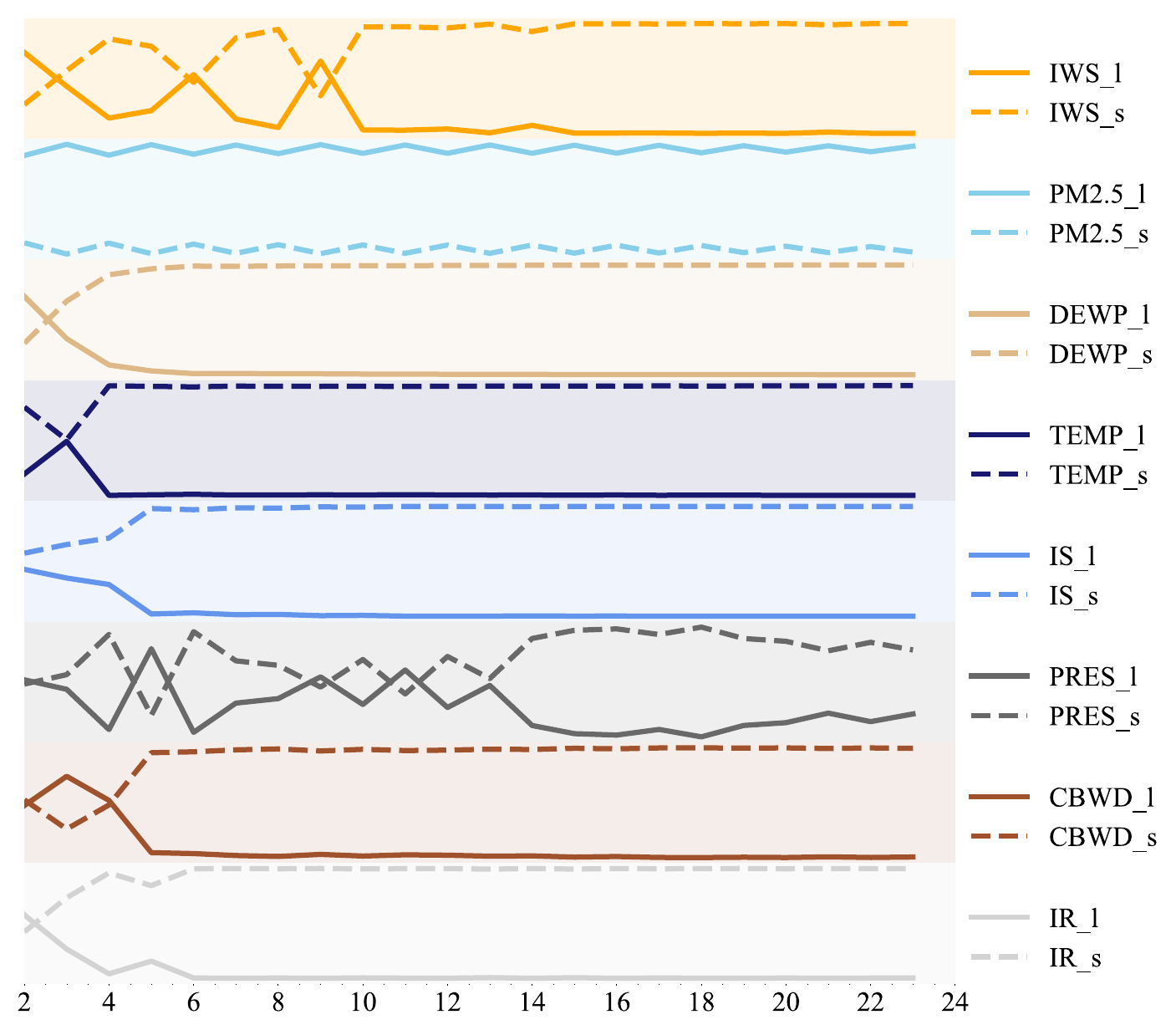}
	\caption{PM2.5 Long (l)/Short (s) term effects evolving over time (24h).}
	\label{PM2.5_effect}
\end{figure}

\normalsize
PM2.5 is fine inhalable particles, with diameters that are generally 2.5 
micrometers and smaller. Exposure to such fine 
particles has been linked to early death from heart and lung disease 
\cite{franklin2008role}. Understanding influential variables and forecasting 
PM2.5 accurately are necessary so that people can avoid going outside on time 
with high PM2.5 levels. We report the long-term and short-term effects  of each 
variable during the day in Figure \ref{PM2.5_effect}.

As shown in Figure \ref{PM2.5_effect}, the impact of prior PM2.5 information is 
more significant than that of recent fresh information for real-time 
forecasting PM2.5 in a day. It indicates that PM2.5 always has long-term 
effects; we should utilize both past and current data of PM2.5. On the other 
hand, for other variables, such as dew point (DEWP), hours 
of snow(IS), temperature (TEMP), and the hours of rain (IR), the past information is only helpful 
before 6:00 am. After that, the ratio of instantaneous influence occupies 
almost $99\%$ for 
predicting PM2.5 in the next hour, showing little long-term effect. Therefore, 
we can conclude that it is sufficient to see the current data of these 
variables for forecasting PM2.5 during the daytime. One possible 
reason is that the natural environment has a significant impact on PM2.5 
changes when there is less human activity and less pollution released at night, 
but human activity plays an important role during the day, so the long-term 
effects of these meteorological variables are lost.

\begin{figure}[ht]
	\centering
	\includegraphics[scale= 0.38]{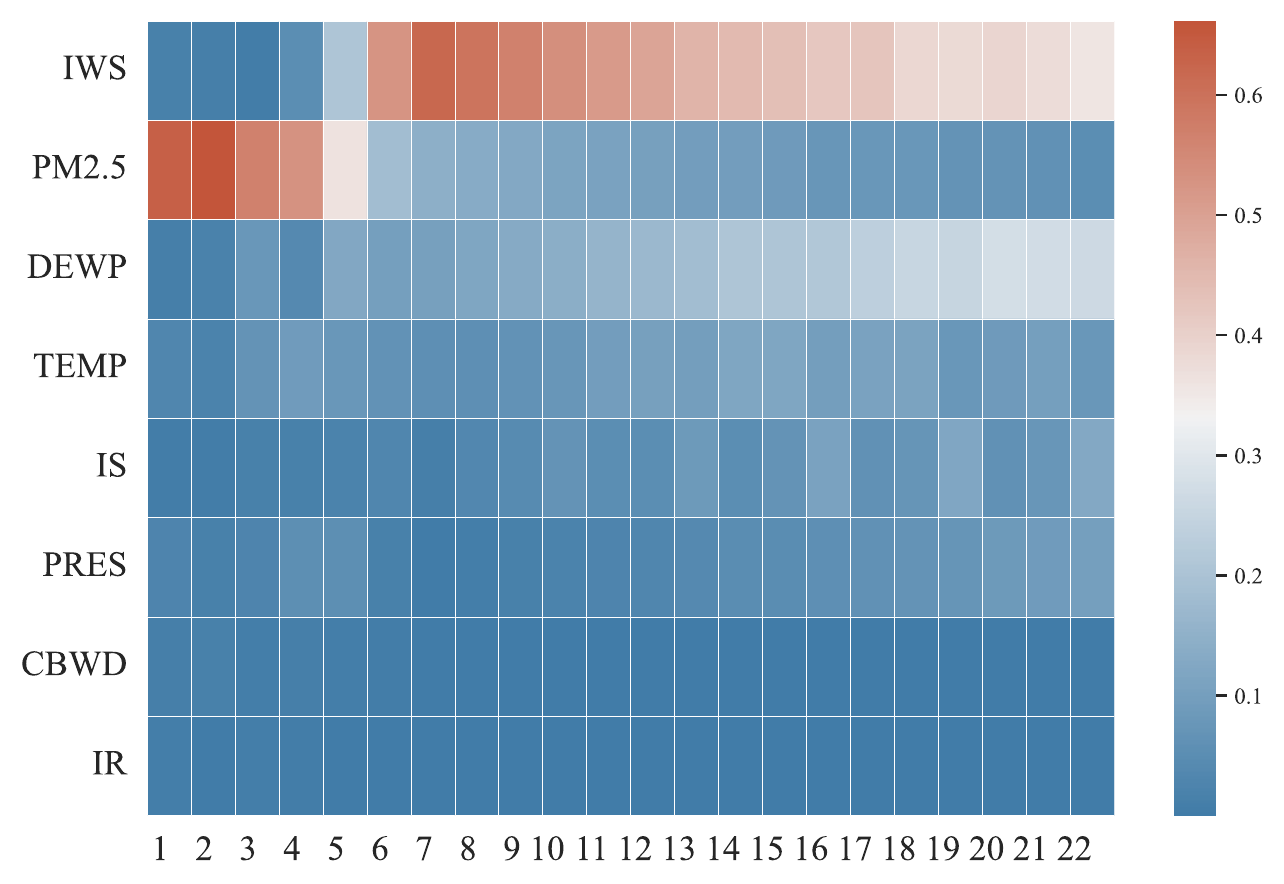}
	\caption{PM2.5 variable importance.}
	\label{PM2.5_variable}
\end{figure}

Figure \ref{PM2.5_variable} shows the dynamic change of variable importance 
considering the long-term effect and instantaneous influence.
As shown in Figure \ref{PM2.5_variable}, in the early morning, the PM2.5 itself 
contributes more to predict PM2.5 in the next hour. In the daytime, the wind 
speed, dew point, temperature, all take on greater significance.
Taking into account all time steps, the top four important variables are the 
wind speed, PM2.5, the dew point, and the temperature. 
According to \cite{pu2011effect}, wind speed has a significant impact on the amount of such inhalable particles transported and dispersion between 
Beijing and its surrounding areas. In addition, \cite{liang2015assessing} also conclude that dew point and temperature are critical factors for PM2.5 prediction. 
Therefore, our variable importance is in line with the domain knowledge.

\subsubsection{Case Study III: Exchange Data}
\begin{figure}[ht]
	\centering
	\includegraphics[scale=0.343]{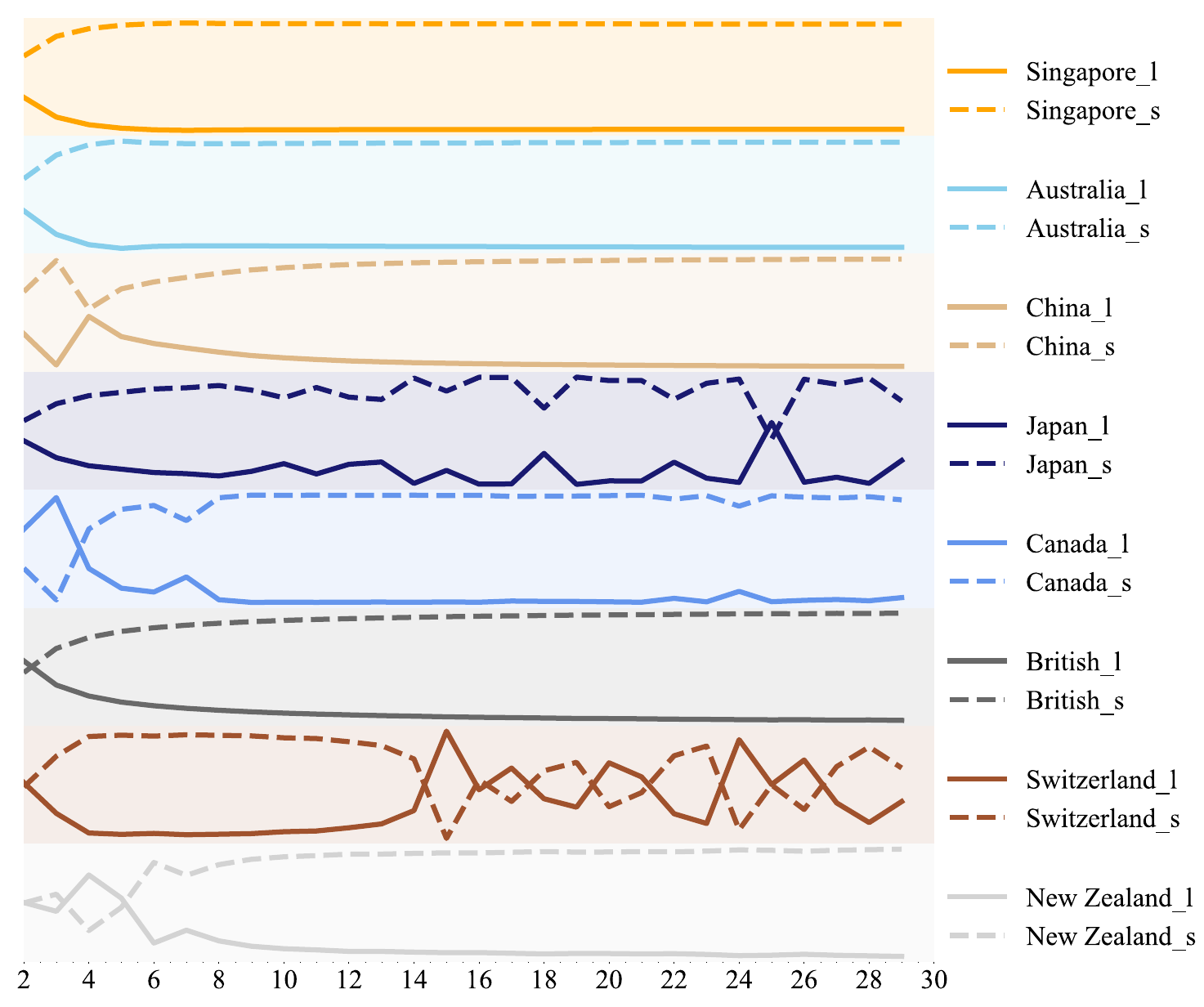}
	\caption{Exchange  Long (l)/Short (s) term effects evolving over time (30 days).}
	\label{exchange_effect}
\end{figure}

\normalsize
Investors need to be very aware of changes in foreign exchange rates. These 
changes greatly impact the returns on foreign investments. As a result, if the 
exchange rate can be predicted with accuracy, investors could 
improve the timing of their foreign investments and earn higher returns. The long-term and short-term effects evolving over time for predicting Singapore’s exchange rate on the next day is shown in Figure \ref{exchange_effect}. The 
exchange rates of Japan and Switzerland have both an immediate and long-term 
effects. Nevertheless, for other countries, 
the instantaneous influence is gradually increasing compared with the long-term 
effect.

\begin{figure}[ht]
	\centering
	\includegraphics[scale= 
	0.38]{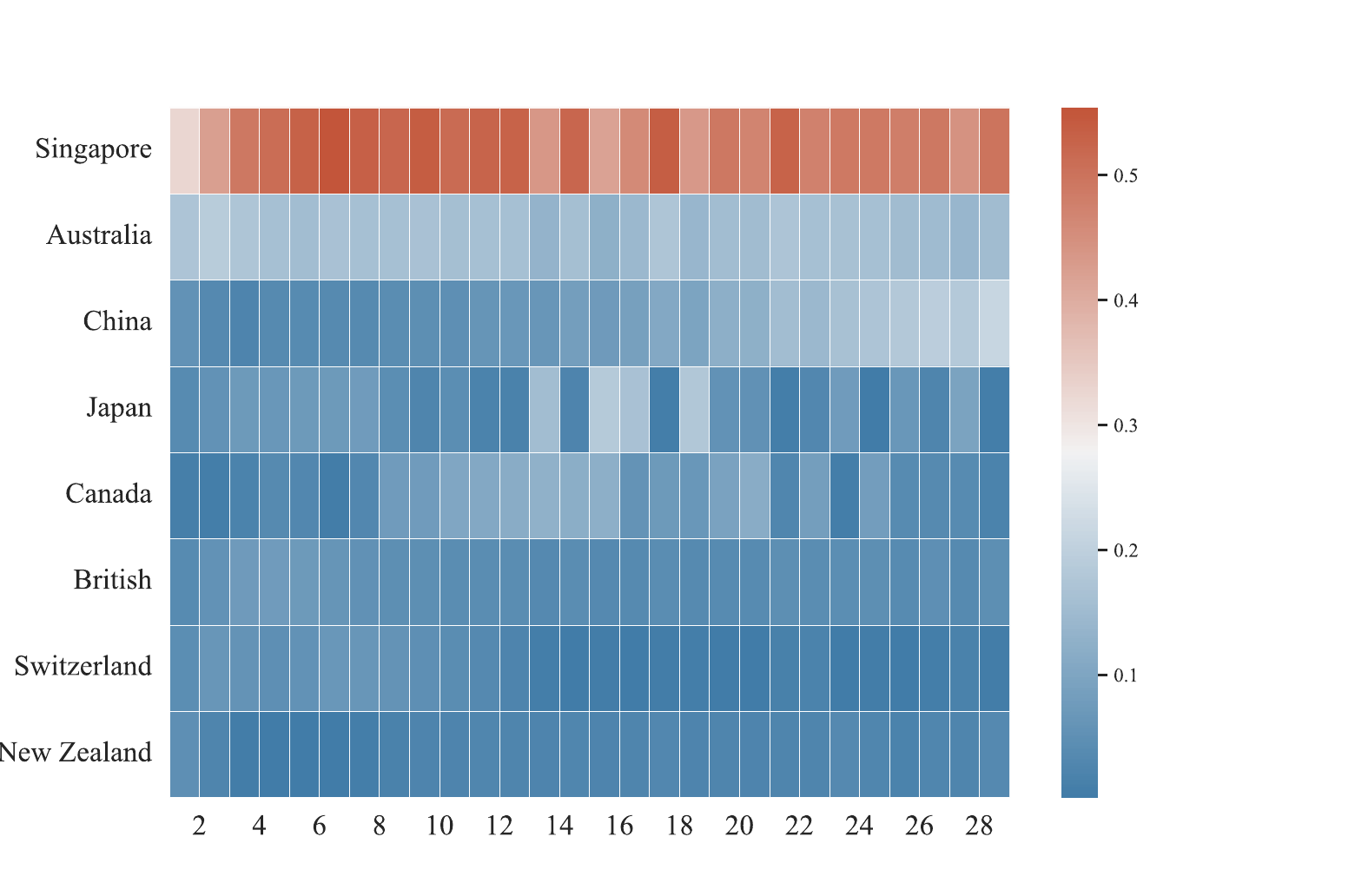}
	\caption{Exchange variable importance.}
	\label{exchange_variable}
\end{figure}

Figure \ref{exchange_variable} displays the variable importance changing over 
time. Singapore has the biggest influence on predicting Singapore’s exchange rate on the next day, 
followed by Australia, China, and Japan. 

In summary, three case studies demonstrate that our proposed DeLELSTM can not 
only provide a transparent and clear explanation but also be able to 
distinguish the instantaneous influence and long-term effect of each variable. 
These explanations aid us in better understanding the data and building models 
for time series prediction.

\subsection {Baseline Comparisons}
In this subsection, we compare the interpretation results of the three baseline 
models, i.e., RETAIN, IMV-Full, and IMV-Tensor. Due to the page limitation, we 
show the variable importance at each time of three 
models in the supplementary. 

As shown in Figure 1,2,3  in the supplementary\textsuperscript{\ref{supp}}, we can observe that the interpretable models based on the attention 
mechanism give the variable importance always the same at most of the time 
points. In addition, except for the most important variables identified by the 
models, it is hard for them to distinguish the importance of the other 
variables. The reason might be the hidden state representations are similar 
across time steps \cite{mohankumar2020towards}.

To further evaluate the sufficiency of our explanation model, we retain the top 50\% of features identified by each explainable model, and then feed them into the LSTM of the same architecture to obtain the prediction results. Table \ref{compare} shows the prediction results on the electricity consumption task. We can observe that features selected by our model can obtain better performance. 

\begin{table}
\centering
  	\begin{tabular}{lccc}
		\hline
		Model      & RMSE          & MAE           & MAPE            \\
		\hline
		DeLELSTM   & \textbf{1.677±0.022} & \textbf{1.116±0.019} & \textbf{1.746\%±0.038\%} \\
		IMV-Full   & 5.332±1.232 & 3.964±0.959 & 6.264\%±1.418\% \\
		IMV-Tensor & 1.686±0.020 & 1.130±0.022 & 1.776\%±0.046\% \\
		RETAIN     & 1.695±0.027 & 1.146±0.025 & 1.807\%±0.046\%\\
		\hline
	\end{tabular}
	\caption{Performance based on top 50\% important variables}
	\label{compare}
\end{table}
Compared with these baseline models, our proposed model is able to identify the 
dynamic impacts of variables on the prediction over time, and distinguish the 
instantaneous influence and the long-term effects of each variable.

\section{Conclusion}
Explaining the time series forecasting model is of significance, especially in 
high-stakes applications. In this work, we propose DeLELSTM, a 
decomposition-based linear explainable LSTM, to improve the interpretability of 
LSTM. Specifically, DeLELSTM decomposes the hidden states into the linear 
combination of the past information and the new information of each variable so 
that it can capture the instantaneous influence and long-term effects. The 
utilization of linear regression also guarantees that the explanations are 
transparent and clear. The experimental results on three real datasets 
demonstrate the effectiveness of DeLELSTM compared with baseline models. The 
case studies show that the explanations made by DeLELSTM are in line with 
domain knowledge. 

One limitation of this work is that the contributions to the prediction are for 
individual variables, ignoring the complex interactions between variables, 
which is common in real life. We keep this challenging task as our future work. 

\section*{Acknowledgments}
The work described in this paper was partially supported by the InnoHK initiative, The Government of the HKSAR, and the Laboratory for AI-Powered Financial Technologies. Qi WU acknowledges the support from The CityU-JD Digits Joint Laboratory in Financial Technology and Engineering; The Hong Kong Research Grants Council [General Research Fund 14206117, 11219420, and 11200219]; The CityU SRG-Fd fund 7005300, and The HK Institute of Data Science.

\bibliographystyle{unsrt}  
\bibliography{references}

\end{document}